\documentclass[journal]{IEEEtran}
\usepackage[utf8]{inputenc}

\usepackage[style=ieee]{biblatex}
\usepackage{graphicx}
\usepackage{array}
\usepackage{bm}
\usepackage{amsmath}
\usepackage{amssymb}
\usepackage{tabularx}

\usepackage{textcomp}
\usepackage{gensymb}
\usepackage{overpic}
\usepackage{xcolor}
\usepackage{lipsum}
\usepackage{color}
\begin{document}

\title{Model Predictive Control for Human-Centred Lower Limb Robotic Assistance} 
\author{Christopher~Caulcrick, Weiguang~Huo, Enrico~Franco, Samer~Mohammed, Will~Hoult, and Ravi~Vaidyanathan%
\thanks{C. Caulcrick, W. Huo, E. Franco, and R. Vaidyanathan are with the Department of Mechanical Engineering, Imperial College London, London, UK. (e-mail: christopher.caulcrick12@imperial.ac.uk; w.huo@imperial.ac.uk; e.franco11@imperial.ac.uk; r.vaidyanathan@imperial.ac.uk).}%
\thanks{S. Mohammed is with the LISSI Lab, University of Paris-Est Cr\'{e}teil, 94400 Vitry-sur-Seine, France. (e-mail: samer.mohammed@u-pec.fr).}%
\thanks{W. Hoult is with McLaren Applied, McLaren Technology Centre, Woking, UK. (e-mail: will.hoult@mclaren.com)}}

\markboth{}%
{Caulcrick \MakeLowercase{\textit{et al.}}: Model Predictive Control for Human-Centred Lower Limb Robotic Assistance}

\maketitle
\begin{abstract}

Loss of mobility or balance resulting from neural trauma is a critical consideration in public health. Robotic exoskeletons hold great potential for rehabilitation and assisted movement, yet optimal assist-as-needed (AAN) control remains unresolved given pathological variance among patients. We introduce a model predictive control (MPC) architecture for lower limb exoskeletons centred around a fuzzy logic algorithm (FLA) identifying modes of assistance based on human involvement. Assistance modes are: 1) passive for human relaxed and robot dominant, 2) active-assist for human cooperation with the task, and 3) safety in the case of human resistance to the robot. Human torque is estimated from electromyography (EMG) signals prior to joint motions, enabling advanced prediction of torque by the MPC and selection of assistance mode by the FLA. The controller is demonstrated in hardware with three subjects on a 1-DOF knee exoskeleton tracking a sinusoidal trajectory with human relaxed, assistive, and resistive. Experimental results show quick and appropriate transfers among the assistance modes and satisfied assistive performance in each mode. Results illustrate an objective approach to lower limb robotic assistance through on-the-fly transition between modes of movement, providing a new level of human-robot synergy for mobility assist and rehabilitation.


\end{abstract}


\begin{IEEEkeywords}
Model predictive control, exoskeleton, assist-as-needed, fuzzy logic, human-robot interaction.
\end{IEEEkeywords}

\IEEEpeerreviewmaketitle

\section{Introduction}
\subsection{Problem Context}
\IEEEPARstart{B}{alance} dysfunction resulting from ageing and/or neural trauma is reaching epidemic proportions. In the USA, there are 800,000 new cases of stroke each year \cite{Yang2017}; 40\% of patients suffer serious falls within a year of stroke \cite{Teasell2002}. Indeed, the number of falls per year due to stroke is expected to reach 500,000 in the USA alone by 2040, representing a total annual cost of \$16 billion per year \cite{Burge2007}. Population-based studies have shown a 35\% prevalence of gait disorders among persons over age 70, and 80\% over 85 years of age \cite{Verghese2006}. The UK societal cost of stroke exceeds \pounds8 billion a year \cite{Saka2009}. Following acute medical response, long-term patient outcomes are almost entirely dependent on rehabilitation and early initiation is critical. Outcome can be impacted by beginning activity in as little as 24 hours after stroke trauma \cite{Indredavik1999} and there is broad agreement that regular activity within the first 6 months of incidence is vital for best possible recovery \cite{Kwakkel2004}.

Early activity post-stroke is often impossible without some form of supported movement, yet limited resources mean that most patients receive inadequate rehabilitation; even in-clinic stroke patients are only functionally active 13\% of the day \cite{Gage2007}. Ageing population projections indicate a 44\% increase in strokes in the UK in the next 20 years \cite{Levy2010}. Even partial muscle dysfunction impacts simple independence and activities of daily living. Age-related changes in muscle and joint forces have also been demonstrated in movements such as sit-to-stand \cite{Smith2020} which demands up to 97\% of available muscle strength in the elderly \cite{Hughes1996}. Stroke patients consequently take significantly longer to execute simple movements, entailing greater risk of falls and poorer capacity to function \cite{Mao2018}. The importance of lower limb rehabilitation for mobility and its role in increased risk of social isolation have led to a range of rehabilitation therapies; however, exploiting the full potential of assistive technologies in clinical settings remains challenging. Telemedicine in rehabilitation is an area of vital future importance \cite{Atashzar2012,Burridge2017}. Neural trauma and elderly patients are in firm need of tools to support their balance, mobility and rehabilitation out-of-clinic.

Wearable robots – i.e. robotic exoskeletons – offer the potential for supported movement and immediate therapy post trauma which is critical for neural rehabilitation as well as general mobility. Compared with conventional therapy, robotic rehabilitation may deliver highly controlled repetitive and intensive training, reduce the burden of clinical staff as well as providing a quantitative assessment of motion and forces \cite{Dollar2008,Huo2016,Young2017}. Stationary devices such as GaitTrainer \cite{Hesse2000} and HapticWalker \cite{Schmidt2007} report both patient cognitive engagement and compliance during locomotor training. Furthermore, studies using wearable exoskeletons such as Hybrid Assistive Limb \cite{Maeshima2011}, Ekso \cite{Strausser2010} and H2 \cite{Bortole2015} have argued that relevant sensory inputs and central neuronal circuits become activated under physiological conditions (i.e. overground walking) leading to neural regeneration. Significant effort has been focused on synergising exoskeleton action for patients with the capacity to provide partial muscular efforts (e.g. elderly or stroke patients), for whom the exoskeleton complements the wearer’s strength to complete a successful movement. This demands the combination of augmenting mobility, lowering risk of injury and, potentially, supporting rehabilitation. The majority of patient beneficiaries are at serious risk of falls from loss of control \cite{Riley1997} as well as sustained damage due to unhealthy compensation \cite{Smith2020}. However, this group also has the potential of rehabilitation if exoskeleton control can synergise its action with user muscle activity. Achieving this demands an exoskeleton to be intelligent enough to coordinate its activity with human muscle action, recognise and reward human effort to maximise rehabilitation, and be compliant such that it does not inhibit user movements.

\subsection{Related Work}
Intensive rehabilitation training and exercises are vital to help stroke patients relearn patterns of motion, enabling them to move again \cite{Srivastava2015}; exercises must be appropriately adjusted according to the patients' rehabilitation stages. There are three common types of these exercises: passive, active-assist, and active, which are respectively used in the early, middle and late stages of stroke rehabilitation. Passive exercises completely rely on the assistance of therapists; active-assist exercises are used when patients have limited mobility but still need therapists' assistance for performing a full range of motion; active exercises can take place without help from therapists once voluntary muscle control is recovered \cite{Kahn2006}. Such subject-specific rehabilitation processes are largely labour-intensive and costly. Understandably, development of robots for assisting post-stroke patients in daily living activities and rehabilitation training is recently attracting great attention \cite{Chang2013}.

Robotic exoskeletons have typically been implemented with focus on one type of exercise \cite{Huang2019,Huang2015,Wu2018,Huo2018a} and position control has been used to carry out passive exercise. For active-assist, robotic devices are required to provide appropriate assistance by sensing patient participation and/or understanding the patient's physical limitations, much like a trained therapist. Control approaches delivering active-assist have been termed assist-as-needed (AAN) \cite{Cai2006,Shahbazi2016,Shahbazi2018,Atashzar2017}, i.e. allowing patients to complete movement without suppressing effort contribution. Although studies have shown AAN control can support the recovery of stroke patients, finding an optimal general AAN algorithm is challenging because the pathology can vary greatly with each case, and the rate of recovery is highly unpredictable \cite{Emken2007}.

A common solution for providing active-assist to the impaired limb is to increase compliance of passive servo-type controllers. This can be achieved either using mechanically compliant joints \cite{Hussain2017} or by reducing control gains \cite{Garrido2014}. The main advantage of such approaches is that it is relatively simple to implement as it does not require knowledge of the underlying system model. However, they are not adaptive to individual condition and have limited effectiveness for stimulating neuroplasticity as patient effort is not actively encouraged \cite{Levi2014}. Meanwhile, impedance control is employed for many assistive exoskeletons, aiming to exhibit human-like mechanical properties and implement a good compromise between tracking ability and compliance \cite{Huo2020,Proietti2016}. Conventional impedance control may require trial-by-trial adaptation to obtain suitable impedance, which varies from subject to subject and with muscular fatigue. Applying such methods in clinical situations is therefore not straightforward, particularly for early stages of recovery. In this context, a number of adaptive impedance approaches have been developed to promote and maintain patient participation \cite{Khan2015,Perez-ibarra2015}. P\'{e}rez-Ibarra et al. \cite{Perez-ibarra2019} proposed an assistive-resistive strategy, which can determine appropriate task difficulty by estimating the dynamic contribution of the patient based on an exoskeleton dynamic model. However, drawbacks of such model-based methods include latency in human effort estimation and sensitivity to external disturbances. A small body of work has explored offline adaptive control to periodically change task difficulty, i.e. adapting trial-by-trial to modulate control parameters based on predesigned performance indexes. For example, Balasubramanian et al. \cite{Balasubramanian2008} designed a wearable upper-extremity robot for functional therapy in repetitive activities of daily living (ADL). An iterative learning control was used to tune the assistive torque trial-by-trial. Such approaches also face a latency problem, not responding quickly to changes of the wearer's motion intention and rehabilitation requirements.

Model predictive control (MPC), which is considered an effective method for addressing the latency problem, has been investigated for AAN rehabilitation previously. Ozen et al. applied nonlinear MPC to an upper body rehabilitation virtual task with a delta robot, optimizing for training task completion, motor learning, and trainee skill level simultaneously \cite{Ozen2019}. This approach is promising for neurorehabilitation, however it does not consider human joint behaviour directly. Raza et al. explored using MPC to control a simulated upper limb rehabilitation robot under disturbance conditions. Task precision improved however the approach was unable to distinguish external disturbance from human effort \cite{Raza2018}. Teramae et al. applied MPC to a 1-DOF upper limb rehabilitation task by considering an upper-limb exoskeleton model and using muscle activity for human joint torque estimation \cite{Teramae2018}. The method was able to derive optimal robot torque for completing the task, but did not adapt the objective function for different assistive modes required during rehabilitation training, particularly safety mode in the case of human resistance to the robot.

\subsection{Aims and Objectives}

In this paper, an AAN control framework is proposed which combines MPC with a fuzzy logic-based algorithm (FLA) to deliver optimal robotic assistance based on the human-robot interaction. The main contribution of this paper lies in 1) design of a control framework for three important rehabilitation modes: passive, active-assist, and safety; 2) development of a fuzzy logic-based MPC algorithm for choosing appropriate assistive modes on-the-fly, meanwhile, providing optimal power assistance by taking into account the wearer's effort; and 3) the experimental studies for validating the proposed control approach. The nature of assistance is preemptively adapted based on wearer's muscle activities (electromyography (EMG) signals), which are generated prior to the corresponding human motions \cite{Winter2009}. MPC is chosen to provide optimal torque assistance over a prediction horizon, while taking advantage of the intention prediction capability of EMG signals. The FLA estimates the human-robot interaction modes using reference angular velocity and human joint torque estimated from EMG signals. The likelihood, or suitability, of three assistance modes, passive, active-assist, and safety, are directly integrated into the MPC by tuning the weighting of costs in the MPC objective function. The use of MPC also allows us to perform constrained optimization, which can take into consideration state limitations for subject safety and hardware protection. The proposed method is demonstrated effectively using a single-joint knee exoskeleton with three healthy subjects in three human-robot interaction modes. In passive mode, the results show that the exoskeleton is dominant and the human-exoskeleton system closely tracks the target trajectory. In active-assist mode, the exoskeleton can enable wearers to contribute productive effort to the desired tasks. The level of compliance and human effort contribution can both be tuned using a single parameter in the controller. During the safety mode, the exoskeleton is highly compliant, permitting substantial tracking error, when the wearer is resistive to the exercise.

The rest of this paper is organised as follows. Section \ref{sec_method} describes the method, which includes a system overview, details of the MPC, design of the FLA, and description of the human torque estimation (HTE) model. The experimental setup is described in Section \ref{sec_experimental_setup}, including details of the exoskeleton, EMG signal acquisition, the process of parametric identification, analysis of controller response to human torque, and experimental protocol. In Section \ref{sec_results} results are presented from an experimental verification of our proposed AAN control method for knee joint flexion/extension in a sitting position with three healthy subjects. Finally, a conclusion is presented in Section \ref{sec_conclusion}.
\begin{figure}[!t]
    \centering
    \begin{overpic}[width=0.9\columnwidth]{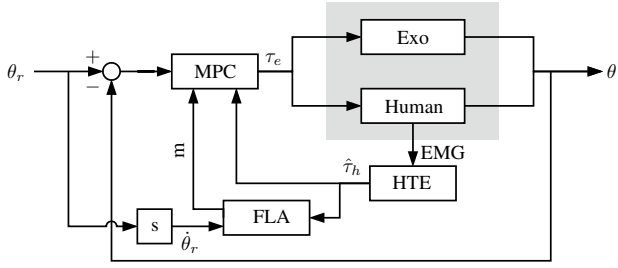} 
        \put(-2,33.25){\scalebox{0.75}{$\theta_{r}$}}
        \put(98,33.25){\scalebox{0.75}{$\theta$}}
        \put(41,36){\scalebox{0.75}{$\tau_e$}}
        \put(11,36){\scalebox{0.75}{$+$}}
        \put(11,31){\scalebox{0.75}{$-$}}
        \put(29.25,33.25){\scalebox{0.75}{MPC}}
        \put(63,39){\scalebox{0.75}{Exo}}
        \put(61,27.5){\scalebox{0.75}{Human}}
        \put(62.25,14.75){\scalebox{0.75}{HTE}}
        \put(67,19.75){\scalebox{0.75}{EMG}}
        \put(54,17.5){\scalebox{0.75}{$\hat{\tau}_h$}}
        \put(22,7.75){\scalebox{0.75}{s}}
        \put(27,4.5){\scalebox{0.75}{$\dot{\theta}_r$}}
        \put(39,9){\scalebox{0.75}{FLA}}
        \put(26,20){\scalebox{0.75}{\rotatebox{90}{m}}}
    \end{overpic}
    \caption{Schematic showing control framework with model predictive controller (MPC), plant (Exo and Human), model for human torque estimation (HTE) from EMG signals, and fuzzy logic-based algorithm (FLA). $\theta$ and $\theta_r$ are the knee joint angle and reference joint angle, $\dot{\theta}_r$ is the reference joint velocity, $\tau_e$ and $\hat{\tau}_h$ are exoskeleton torque and estimated human torque, and $m$ is the assistance mode variable.}
    \label{fig:controller}
\end{figure}

\section{Method} \label{sec_method}

The AAN controller proposed in this work combines MPC for trajectory tracking with fuzzy logic for determining the correct assistance mode based on the human involvement. The control framework structure is depicted in Fig.~\ref{fig:controller}. The MPC generates optimal exoskeleton torque by taking into account reference joint angle, estimated human torque, and inferred appropriate mode of assistance.
EMG signals are used to estimate the human joint torque using the HTE model, and the estimated human torque is used by the FLA, with the reference joint velocity to generate the assistance mode.

\subsection{Human-Exoskeleton System} \label{sec:system}
\begin{figure}
    \centering
    \begin{overpic}[width=1.0
    \columnwidth]{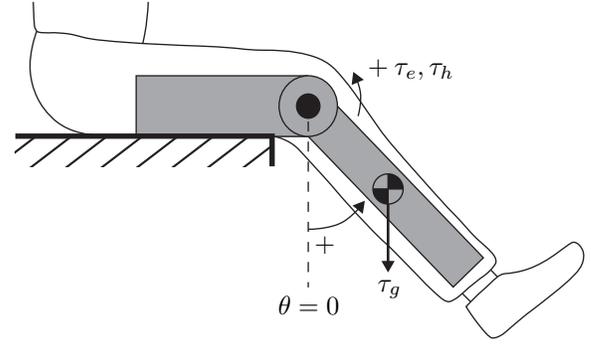} 
        \put(48.5,5){\scalebox{1}{$\theta=0$}}
        \put(54,14.5){\scalebox{1}{$+$}}
        \put(62,41){\scalebox{1}{$+\:\tau_e,\tau_h$}}
        \put(63.5,8.5){\scalebox{1}{$\tau_g$}}
    \end{overpic}
    \caption{Schematic showing human-exoskeleton system in the sitting position, $\theta$, joint angle, $\tau_e$ exoskeleton torque, $\tau_h$, human torque, and $\tau_g$, gravitational torque of the shank/foot segment.}
    \label{fig:hebi-exo_schematic}
\end{figure}

For a 1-DOF lower limb exoskeleton assisting knee flexion and extension in a planar seated position (see Fig. \ref{fig:hebi-exo_schematic}), a model can be written as
\begin{equation}
    J\ddot{\theta}+B\dot{\theta}+\tau_g\sin\theta - \tau_h  = \tau_e
    \label{eqn:model}
\end{equation}
where $\theta$ is the knee joint angle, $J$ is the lumped inertia of human and exoskeleton shank-foot, $B$ is the lumped viscous friction parameter, $\tau_g$ is the lumped gravitational torque, and $\tau_e$ and $\tau_h$ are the exoskeleton torque and human torque respectively. Note that the model assumes the exoskeleton and wearer's joints are rigidly coupled so knee extension/flexion movements are considered synchronous and simultaneous \cite{Huo2018}. Model parameters $J, B$ and $\tau_g$ are not known but are identified experimentally in Section \ref{sec:param_id}.

\subsection{Model Predictive Controller}

As shown in Fig.~\ref{fig:controller}, the MPC approach provides constrained nonlinear optimal control for the human-exoskeleton according to the human effort, tracking error, and assistive mode. The optimisation of the MPC is shown as

\begin{subequations}
\begin{align}\label{eq.MPC}
    \underset{u}{\text{min}} &\quad J = \sum_{k=t}^{t+T-1} (m_k w_\theta(\theta_{r, k}-\theta_{k})^2+w_{\tau}u_{k}^2)\\
    s.t. &\quad \ddot{\theta}_{k+1}= \frac{1}{J}(B\dot{\theta}_k+\tau_g\sin\theta_k - \hat{\tau}_{h,k} - \tau_{e,k}) \\
    &\quad \theta\in[\theta_\text{min} \,\, \theta_\text{max}], \quad \dot{\theta}\in[\dot{\theta}_\text{min} \,\, \dot{\theta}_\text{max}]\\
    &\quad \tau_e\in[\tau_{e,\text{min}}\,\,\tau_{e,\text{max}}]
\end{align}
\end{subequations}
where $T$ shows the prediction horizon, and $k$ denotes the $k$th sample. The prediction horizon can be chosen to reflect the length of time interval from EMG signal activation to joint torque output \cite{Vos1991}. $w_\theta$ is the tracking weight, affecting tracking error cost, and $w_\tau$ is the torque weight, affecting the robot torque cost. $m$ is the mode function which changes robot behaviour by scaling $w_\theta$. The value of $m\in [0\quad1]$ is determined by the FLA (see section {\ref{sec:fla}}) which uses inputs of the reference angular velocity, $\dot{\theta}_r$, and the estimated human torque, $\hat{\tau}_h$, evaluated using the HTE model described in section \ref{sec_tau_est}. 

The cost function is solved in real-time using nonlinear MPC software GRAMPC, suitable for sampling times in the (sub)millisecond range. It uses an algorithm based on an augmented Lagrangian formulation with a tailored gradient descent method for the inner minimisation problem \cite{Kapernick2014}.

A benefit of this MPC approach is the ability to find optimal solutions while applying constraints to the state variables and control input by adjoining the constraint equations to the objective function. Constrained optimisation inherently limits the range and velocity of joint motion to prevent subject discomfort and protect hardware. Constraints of minimum and maximum joint angle, $\theta\in[0\quad1.4]$, angular velocity, $\dot{\theta}\in[-2\quad2]$, and control input torque, $\tau_e\in[-25\quad25]$, are applied. The constraints on joint states can potentially be used by clinical practitioners to personalise the control framework for the patient based on recovery progress and physical limitations.

\subsection{Fuzzy Logic Algorithm} \label{sec:fla}

The assistance mode function, $m\in$ [0 1], adapts the nature of human-robot interaction (see eq.(\ref{eq.MPC})). It can change the robot's effective compliance to allow human contribution to the achievement of the rehabilitation task. When $m = 0$, tracking weight is eliminated entirely and the system operates with the human dominant. It allows significant tracking error and shows maximum compliance. When $m = 1$, tracking weight is the maximum (high gain) value and the system operates with the robot dominant. It shows minimal compliance and limited tracking error. The value of $m$, and therefore nature of assistance on the spectrum of human/robot dominance, is determined using fuzzy logic. Estimated human torque, $\hat{\tau}_h$, and reference angular velocity, $\dot{\theta}_r$, are the FLA inputs.

Given the controller model can fully compensate for the gravity of the exoskeleton and human shank-foot, when estimated human torque, $\hat{\tau}_h$, and reference angular velocity, $\dot{\theta}_r$, are acting in the same direction, the subject is considered to be cooperating with the task. The FLA then identifies that the exoskeleton should operate in active-assist mode (A). On the other hand, with significant estimated human torque acting against the direction of reference angular velocity, the FLA identifies that the exoskeleton should operate in safety mode (S). When the estimated human joint torque is close to 0, then, the human is assumed in the passive mode (P). The likelihood of active-assist and safety mode of operation being suitable for the human involvement condition are defined as, $\mu_A \in [0 \quad 1]$ and $\mu_S \in [0\quad 1]$, respectively. Given the aforementioned analysis, the fuzzy logic rules are designed as shown in Table~\ref{tab:rules}. The estimated human joint torques are classed into two levels, ``Positive" and ``Negative", while the reference joint angular velocity is divided into three levels, ``Positive", ``Zero" (referred to as ``Non Negative" in Table~\ref{tab:rules}), and ``Negative".

Leading up to the implementation of the fuzzy logic rules, a symmetrical Gaussian function is used as the membership function for defining the ``Zero" value for the reference angular velocity, which is given by

\begin{equation}
    f(x) = e^{\frac{-(x-c)^2}{2\sigma^2}}
\end{equation}
where $x$ is the input and parameters $\sigma$ and $c$ determine the width and centre of the Gaussian distribution respectively. The distribution centre, $c$, is located about zero and the standard deviation, $\sigma$, is small. For evaluating the ``Negative" and ``Positive" values for estimated human joint torque and reference velocity, a sigmoidal membership function is used. The sigmoidal function can be written as
\begin{equation}
    f(x) = \frac{1}{1 + e^{-a(x-c)}}
\end{equation}
where $a$ and $c$ are function parameters affecting steepness and centre respectively. Both membership functions have the benefit of being continuous and smooth over the entire range of possible input values. This means changes in controller behaviour do not occur suddenly and jumping between discrete control modes is avoided \cite{Kong2009}. The function parameters are chosen such that membership values cannot sum to greater than one. Membership functions for all inputs are summarised in Table \ref{tab:mfs}.

\begin{figure}
\centering
\begin{overpic}{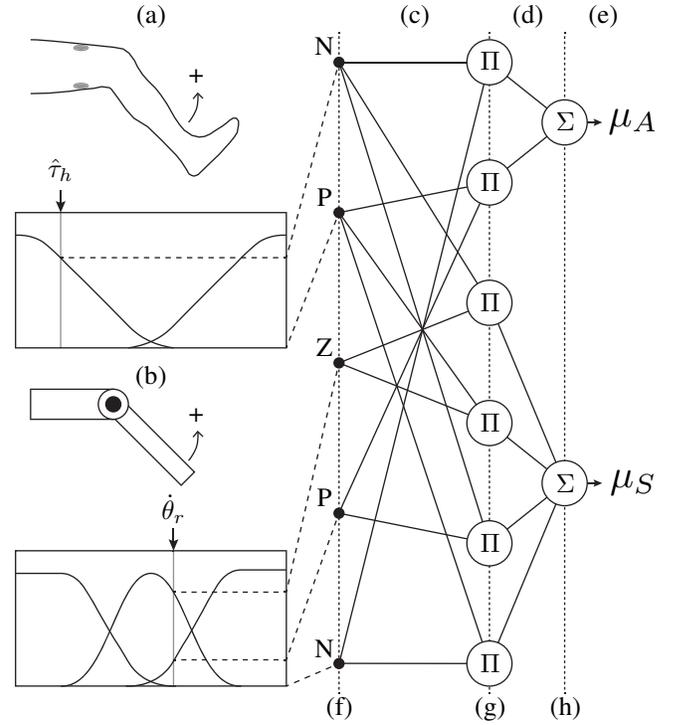} 
    \put(8,75.5){\makebox(0,0){\scalebox{1}{$\hat{\tau}_h$}}}
    \put(26,88){\makebox(0,0){\scalebox{1}{+}}}
    \put(23,30.5){\makebox(0,0){\scalebox{1}{$\dot{\theta}_r$}}}
    \put(26,43){\makebox(0,0){\scalebox{1}{+}}}
    
    \put(43,92){\makebox(0,0){\scalebox{1}{N}}}
    \put(43,72){\makebox(0,0){\scalebox{1}{P}}}
    \put(43,52){\makebox(0,0){\scalebox{1}{Z}}}
    \put(43,32){\makebox(0,0){\scalebox{1}{P}}}
    \put(43,12){\makebox(0,0){\scalebox{1}{N}}}
    
    \put(65,90){\makebox(0,0){\scalebox{1}{$\Pi$}}}
    \put(65,74){\makebox(0,0){\scalebox{1}{$\Pi$}}}
    \put(65,58){\makebox(0,0){\scalebox{1}{$\Pi$}}}
    \put(65,42){\makebox(0,0){\scalebox{1}{$\Pi$}}}
    \put(65,26){\makebox(0,0){\scalebox{1}{$\Pi$}}}    \put(65,10){\makebox(0,0){\scalebox{1}{$\Pi$}}}
    
    \put(75,82){\makebox(0,0){\scalebox{1}{$\Sigma$}}}
    \put(75,34){\makebox(0,0){\scalebox{1}{$\Sigma$}}}
    
    \put(84,82){\makebox(0,0){\scalebox{1.5}{$\mu_A$}}}
    \put(84,34){\makebox(0,0){\scalebox{1.5}{$\mu_S$}}}
    
    \put(20,96){\makebox(0,0){\scalebox{1}{(a)}}}
    \put(20,48){\makebox(0,0){\scalebox{1}{(b)}}}
    \put(55,96){\makebox(0,0){\scalebox{1}{(c)}}}
    \put(70,96){\makebox(0,0){\scalebox{1}{(d)}}}
    \put(80,96){\makebox(0,0){\scalebox{1}{(e)}}}
    \put(45,4){\makebox(0,0){\scalebox{1}{(f)}}}
    \put(65,4){\makebox(0,0){\scalebox{1}{(g)}}}
    \put(75,4){\makebox(0,0){\scalebox{1}{(h)}}}
\end{overpic}
\caption{Fuzzy logic algorithm for inferring assistance mode for human involvement condition. (a) Fuzzy membership functions for human torque. (b) Fuzzy membership functions for reference angular velocity. (c) Fuzzy rule bases (Table \ref{tab:rules}). (d) Rule outputs before summation. (e) Outputs of fuzzy logic. (f) Fuzzy membership values. (g) Inference operators. (h) Summation for assistance modes.}
\label{fig:fla_schematic}
\end{figure}

\begin{figure}[!t]
\centering
\includegraphics[width=\columnwidth]{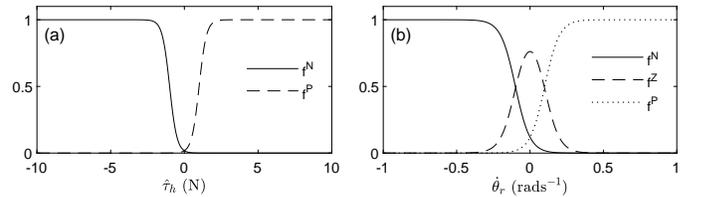}
\caption{Membership functions: a) of estimated human torque, $\hat{\tau}_h$, for negative, $f^{N}(\hat{\tau}_h)$, and positive, $f^{P}(\hat{\tau}_h)$, and b) of the reference angular velocity, $\dot{\theta}_r$, for negative, $f^{N}(\dot{\theta}_r)$, zero, $f^{Z}(\dot{\theta}_r)$, and positive, $f^{P}(\dot{\theta}_r)$.}
\label{fig:mf}
\end{figure}

\begin{table}[!t]
\centering
\caption{Fuzzy Rule Bases for Assistance Mode Selection}
\label{tab:rules}
\begin{tabularx}{\columnwidth}{@{}|c|XX|c|@{}}
\hline

 Rule & $\hat{\tau}_h$ & $\dot{\theta}_r$ & FMV \\ \hline
1 & Negative &   Negative   & $\mu_A \to 1$ \\ \hline
2 & Positive &   Positive   & $\mu_A \to 1$ \\ \hline
3 & Negative &   Non Negative   & $\mu_S \to 1$ \\ \hline
4 & Positive &   Non Positive   & $\mu_S \to 1$ \\ \hline
 
\end{tabularx}
\end{table}
\renewcommand{\arraystretch}{1.0}

\renewcommand{\arraystretch}{1.25}

Fig. \ref{fig:fla_schematic} shows the complete scheme of the proposed fuzzy logic algorithm. Larsen product implication method is used as the inference operator to calculate the likelihoods $\mu_A$ and $\mu_S$ as follows \cite{Martinlarsen1980}:

\begin{align}
\mu_A = & f^{N}(\hat{\tau}_h) \times f^{N}(\dot{\theta}_r)+ f^{P}(\hat{\tau}_h) \times f^{P}(\dot{\theta}_r) \\
     \mu_S = & f^{N}(\hat{\tau}_h) \times(f^{Z}(\dot{\theta}_r)+f^{P}(\dot{\theta}_r)) \\
      & + f^{P}(\hat{\tau}_h) \times (f^{Z}(\dot{\theta}_r)+f^{N}(\dot{\theta}_r))   
\end{align}

where $f$ shows the membership function. The superscript $N$, $Z$, and $P$ denote the ``Negative'', ``Zero'' and ``Positive'', respectively.

For example, for the 3rd rule in Table~\ref{tab:rules}, if $\hat{\tau}_h$ is negative and $\dot{\theta}_r$ is either zero or positive (i.e. $f^{N}(\hat{\tau}_h)\to1$ and ($f^{Z}(\dot{\theta}_r)\to1$ or $f^{P}(\dot{\theta}_r)\to1$)) then safety likelihood tends to one, i.e., $\mu_S\to1$. It should be noted that \emph{when the human is in the relaxed state ($\hat{\tau}_h = 0$), both $\mu_A$ and $\mu_S$ are set to zero} according to the designed membership functions (Fig.~\ref{fig:mf}) and rules (Table~\ref{tab:rules}).

Based on the estimated likelihoods $\mu_A$ and $\mu_S$, the assistance mode function, $m$ is designed as follows:
\begin{equation}
    m = 1 - (p_A\mu_A + p_S\mu_S)
\end{equation}
where $p_A$ and $p_S$ are two constants, with $p_A, p_S \in [0 \quad 1]$. 

Using the assistance mode function, $m$, three assistive modes, passive, active-assist and safety, can be smoothly transitioned between, according to the likelihoods $\mu_A$ and $\mu_S$.  In this study, the parameters are set as $p_A = 0.5$ and $p_S = 1.0$ to reflect the desired level of robot compliance in each operation mode. \emph{In the passive mode, with $\mu_A\to0$ and $\mu_S\to0$, the assistance mode $m$ is set to 1, regardless of settings of $p_A$ and $p_S$}. In this mode, the human-robot interaction is exoskeleton dominant with minimal compliance and tracking error. In active-assist mode, as $\mu_A\to1$, $m\to0.5$ causing the tracking error cost in the MPC cost function to reduce by half, making the robot exhibit compliance and enabling suitable human-robot cooperation during the trajectory tracking task. In safety mode, with $\mu_S\to1$, $m\to0$, causing tracking error cost to be largely eliminated and the robot to show very high compliance. 

In practice the value of the active-assist coefficient $p_A$ could be adjusted by a therapist to vary the ratio of human-to-robot effort exerted during a rehabilitation task. Alternatively, $p_A$ could be adjusted automatically by the controller based on a suitable measurement of subject ability task-by-task. Another example would be scaling of the parameter by a fatigue factor which is a function of time elapsed since beginning of an exercise session.

\renewcommand{\arraystretch}{1.25}

\begin{table}[!t]
\centering
\caption{Details and Parameters of Membership Functions used in the Fuzzy Logic Algorithm}
\label{tab:mfs}
\begin{tabularx}{\columnwidth}{@{}|l|lXXX|@{}}
\hline
Membership~Function            & Type      & $\sigma$ & a   & c    \\ \hline
$f^{Negative}(\hat{\tau}_h)$   & Sigmoidal & N/A    & -4  & -1   \\
$f^{Positive}(\hat{\tau}_h)$   & Sigmoidal & N/A    & 4   & 1    \\
$f^{Negative}(\dot{\theta}_r)$ & Sigmoidal & N/A    & -20 & -0.1 \\
$f^{Zero}(\dot{\theta}_r)$     & Gaussian  & 0.1    & N/A & 0    \\
$f^{Positive}(\dot{\theta}_r)$ & Sigmoidal & N/A    & 20  & 0.1  \\ \hline
\end{tabularx}%
\end{table}

\renewcommand{\arraystretch}{1.0}

\subsection{Human Torque Estimation} \label{sec_tau_est}

A linear agonist-antagonist HTE model is employed for estimating $\tau_h$ based on EMG signals measured from knee flexor and extensor muscles:
\begin{equation}
    \hat{\bm{\tau}}_{h,t+k_1} = \bm{A\textbf{ch}}_t + \bm{b}
\end{equation}
where $\bm{A}$ and $\bm{b}$ are linear parameters and $\bm{\textbf{ch}}_t = [\textrm{ch}_1 \quad \textrm{ch}_2]^T$ is an array of EMG signals, processed using the method described in Section \ref{sec:emg_measurement}. For these experiments, two sensors are used so our model can reduce to the following equation:
\begin{equation}
    \hat{\tau}_{h,t+k_1} = b_0 + a_1 \cdot \textrm{ch}_{1,t} + a_2 \cdot \textrm{ch}_{2,t}
    \label{eqn:torque}
\end{equation}
where $b_0$, $a_1$, and $a_2$ are the linear model parameters to be identified. $k_1$ is a constant which corresponds to the prediction horizon from EMG signals prior to joint movement \cite{Winter2009,Vos1991}. The two EMG sensors are placed on the subject’s vastus medialis ($\textrm{ch}_1$) and biceps femoris ($\textrm{ch}_2$) muscles as they are heavily involved during seated knee extension and flexion. This method is also applicable to other thigh muscles, given both knee flexor and extensor muscles are included, making it versatile and applicable to many rehabilitation robot designs. This estimation approach is sufficiently accurate for this work, given the primary aim is to identify joint torque polarity. The model has the benefit of being simple and therefore straightforward to compute for real-time application \cite{Kiguchi2012}. The HTE model depends on the EMG measurement conditions and the properties of the user's body \cite{Lloyd2003}. Therefore model parameters need to be calibrated for each subject, as in previous studies \cite{Teramae2018}. The parameters are determined experimentally. This process and an evaluation of HTE accuracy are detailed in Section \ref{sec_param_id_tau} which shows that the proposed approach is suitable for robotic rehabilitation.

To evaluate the AAN controller, including MPC, and assistance mode FLA performance, the proposed method was applied to a knee joint tracking task with the following experimental setup.

\section{Experimental Setup} \label{sec_experimental_setup}

\subsection{Exoskeleton System}
\begin{figure}
\centering
\begin{overpic}[width=\columnwidth,grid=false]{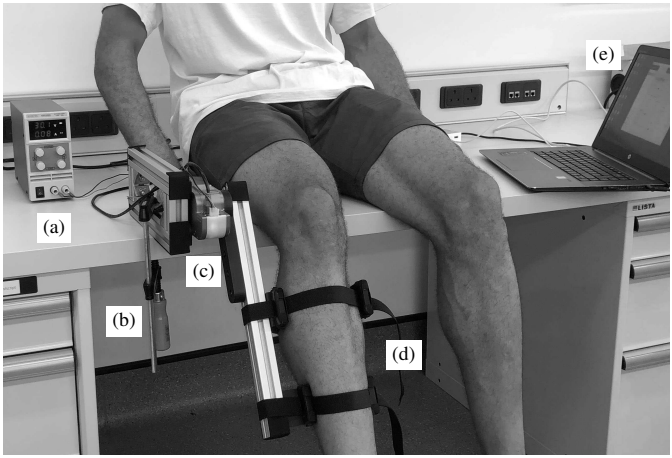}
    \put(7.5, 34){\makebox(0,0){\scalebox{0.75}{\colorbox{white}{(a)}}}}
    \put(18, 20){\makebox(0,0){\scalebox{0.75}{\colorbox{white}{(b)}}}}
    \put(30, 27.5){\makebox(0,0){\scalebox{0.75}{\colorbox{white}{(c)}}}}
    \put(60, 15){\makebox(0,0){\scalebox{0.75}{\colorbox{white}{(d)}}}}
    \put(90, 60){\makebox(0,0){\scalebox{0.75}{\colorbox{white}{(e)}}}}
\end{overpic}
\caption{Experimental setup of the single DOF exoskeleton showing (a) Power supply, (b) F clamp, (c) HEBI X8-9 actuator, (d) Shank segment braces, and (e) Computer with GUI.}
\label{fig:hebi-exo}
\end{figure}

The one Degree-of-Freedom (DOF) knee joint exoskeleton (see Fig. \ref{fig:hebi-exo}) is designed to provide power assistance for rehabilitation purposes. It is made up of two segments, thigh and shank, with the former secured to a stationary surface, and the latter attached to the human shank, coupled with comfortable but secure braces. The braces reduce misalignment between the exoskeleton and human knee joint by adjusting along a rail to fit each subject. Minor changes in the system properties are accounted for in human model parameters. Torque is delivered at the knee joint level by a high power DC series elastic actuator (SEA) (HEBI, USA). To provide a compact and portable structure while giving a high output torque, a compact transmission system is housed inside the actuator unit which is joined directly to the shank and thigh segments. The SEA can safely deliver a maximum torque of 25 Nm. The device is backdrivable and incurs minimal friction cost. Backdrivability is crucial for this application which requires fluid human-exoskeleton torque interaction to facilitate AAN control. Two high resolution incremental encoders are mounted on both motor and joint sides. The output torque of the SEA can be accurately measured based on the rotational deflection between motor-side shaft and output shaft. Angular velocity of the knee joint is obtained by differentiation of joint angle and processed with a low-pass filter (cut-off frequency of 5 Hz) to remove anomalous artefacts from the signal.

The exoskeleton is controlled from a host PC with C++ software integrating the MPC, FLA, HEBI API and Porti amplifier (TMSi, NL) for EMG signal acquisition. The controller and actuator code runs in real-time at 500 Hz, while the EMG amplifier processes and filters signals at 2048 Hz.

\subsection{EMG Measurement} \label{sec:emg_measurement}

\begin{figure}[!t]
\centering
\includegraphics[width=\columnwidth]{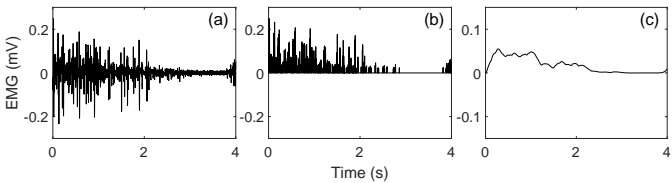}
\caption{EMG processing stages for EMG 1 readings (vastus medialis) during extension assist (EA): (a) raw EMG signals, (b) band-pass filtered between 10 and 500 Hz, noise removed, and full-wave rectified, and (c) low pass filtered for signal envelope at 2 Hz.}
\label{fig:emg}
\end{figure}

Muscle activity signals are captured using two TMSi surface EMG sensors placed on the subject's vastus medialis and biceps femoris muscles. The raw signals are band-pass filtered with corner frequencies of 10 Hz and 500 HZ, full-wave rectified, then lowpass filtered with a cutoff frequency of 2 Hz \cite{Teramae2018,Reaz2006}. This processing extracts the signal envelope which is a good fit for human torque and is used directly in the linear HTE model. Fig. \ref{fig:emg} shows an example of the EMG processing stages during assistive human torque. Processed EMG signals are used to estimate human joint torque over the 0.2 s MPC prediction horizon \cite{Winter2009,Vos1991}.

\begin{table}[!t]
\centering
\caption{Subjects' Information}
\label{tab:subjects}
\begin{tabularx}{\columnwidth}{@{}|l|XXXX|@{}}
\hline
Subject & Sex    & Age & Height (cm) & Weight (kg) \\ \hline
S$_1$   & Male   & 27  & 188        & 85         \\
S$_2$   & Female & 22  & 155        & 53         \\
S$_3$   & Male   & 27  & 179        & 65         \\ \hline
\end{tabularx}
\end{table}

\subsection{Subjects} \label{sec:subjects}
The experiments were conducted with three healthy subjects, able to execute complete flexion and extension movements of the knee joint without spasticity nor contracture. Subject information is provided in Table \ref{tab:subjects}. All subjects were able to actively follow (active-assist mode) or obstruct (safety mode) the knee joint trajectories while observing the progress on a monitor in real-time. The subjects were given a patient information sheet regarding the assessment and gave written informed consent to all experimental procedures. The study received ethical approval from the Research Ethics Committee of Imperial College London. Constraints on the knee joint states and control input were imposed within the controller, and, for safety, device range of motion was securely limited between full extension (90 $\degree$) and full flexion (-30 $\degree$) mechanically. 

\subsection{Parametric Identification} \label{sec:param_id}

Parameters for the seated human-exoskeleton system model (\ref{eqn:model}) and HTE model (\ref{eqn:torque}) require identification. The parameters were identified in three steps.

\subsubsection{Exoskeleton Parameters} \label{sec:param_id_exo}
For this step, the exoskeleton is controlled in the seated configuration, without a human coupled to the system. Sinusoidal reference trajectories, $\theta_r$, with five different frequencies (0.1, 0.2, 0.25, 0.5 and 1.0 Hz) were chosen. The exoskeleton is controlled using a basic proportional controller, i.e. $\tau_e = K(\theta_r - \theta)$, while the knee joint angle $\theta$ is recorded. The joint angular velocity, $\dot{\theta}$, and joint acceleration, $\ddot{\theta}$, are calculated by first and second derivatives of the joint angle low-pass filtered with a cut-off frequency of 5 Hz. From the system model (\ref{eqn:model}), it is clear that $\theta$, $\dot{\theta}$, $\ddot{\theta}$, and $\tau_e$ are known, while the parameters $J_e$, $B_e$, and $\tau_{g,e}$ need to be identified. Given there are the same number of unknown variables as known variables, a linear least-squares optimisation algorithm is used to identify the unknown model parameters \cite{Rifai2017}. Estimated parameter values are shown in Table \ref{tab:params}.

\begin{table}[!t]
\centering
\caption{Human and Exoskeleton System Identified Parameters}
\label{tab:params}
\begin{tabularx}{\columnwidth}{@{}|X|Xll|@{}}
\hline
System  & J ($Kg\cdot m^2$) & B ($Nm\cdot s\cdot rad^{-1}$) & $\tau_g$ ($Nm$) \\ \hline
Exo  & 0.0377                  & 0.0207                                                      & 1.7536                                         \\
S$_1$ & 0.4315                  & 0.1676                                                           & 14.256                                          \\
S$_2$ & 0.1927                  & 0.1534                                                          & 7.5008                                          \\
S$_3$ & 0.3060                  & 0.1575                                                          & 10.595                                         \\ \hline
\end{tabularx}%
\end{table}

\subsubsection{Human Limb Parameters}
The human limb parameters, $J_h$, $B_h$, and $\tau_{g,h}$, can be estimated using the dynamic model (\ref{eqn:model}). Once human-exoskeleton parameters and exoskeleton parameters are determined experimentally, human parameters can be inferred by superposition. Human-exoskeleton system parameters are identified using a similar approach to the exoskeleton parameters, but with human coupled the the exoskeleton. The human subject is instructed to remain relaxed, so human torque can be neglected ($\tau_h = 0$). EMG signals are monitored here to ensure the subject is not exerting any voluntary effort. As in Section \ref{sec:param_id_exo}, the system is proportionally controlled with the same reference trajectories and the unknown human-exoskeleton parameters are identified using linear least-squares. Satisfactory  agreement between input torque, $\tau_e$, and estimated torque, $\hat{\tau}_e$, generated from identified parameters is observed (Average RMSE = 1.47 Nm across all subjects). The estimated human limb parameter values are shown in Table \ref{tab:params}.

\begin{table}[t]
\centering
\caption{Human Torque Estimation Parameters and Error}
\label{tab:tau_h_est}
\newcolumntype{C}{>{\centering\arraybackslash}X}%
\begin{tabularx}{\columnwidth}{|l|CCCc|}
\hline
Subject & $b_0$ & $a_1$ & $a_2$  & RMSE$_{\hat{\tau}_h}$ (Nm) \\ \hline
S$_1$   & 0.181 & 206.2 & -90.5  & 2.96                       \\
S$_2$   & 0.127 & 163.8 & -110.1 & 2.38                       \\
S$_3$   & 0.204 & 181.7 & -132.8 & 3.11                       \\ \hline
\end{tabularx}
\end{table}

\subsubsection{Human Torque Model Parameters} \label{sec_param_id_tau}
\begin{figure}[!t]
\centering
\includegraphics[width=\columnwidth]{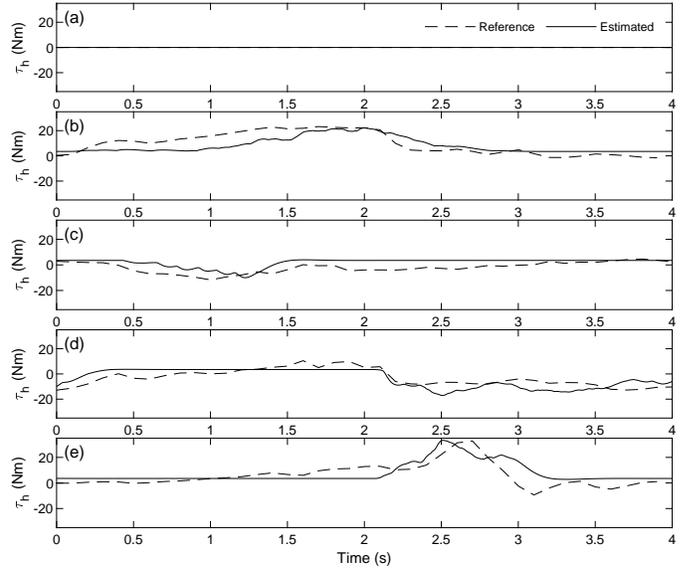}
\caption{Plot showing reference trajectory ($\theta_r$), reference human torque, $\tau_{hr}$, and estimated human torque, $\hat{\tau}_h$, from 4 s of model parameter identification trials for subject 1 with involvement conditions: a) relaxed (R), b) extension assist (EA), c) extension resist (ER), d) flexion assist (FA), and e) flexion resist (FR).}
\label{fig:tau_hs}
\end{figure}

As the combined human-exoskeleton model parameters are already identified and exoskeleton input torque is controlled, the only remaining unknown in the system model (\ref{eqn:model}) is the human torque. For HTE a linear agonist-antagonist model (\ref{eqn:torque}) is used where $b_0$, $a_1$, and $a_2$ are the linear model parameters to be identified. A sinusoidal reference trajectory with frequency of 0.25 Hz was followed using MPC, without human torque included in the system model, and setting $m=1$ in five different involvement conditions: relaxed (R), extension assist (EA), extension resist (ER), flexion assist (FA), and flexion resist (FR). These conditions are representative of the possible human-robot interaction modes for a single knee human-exoskeleton system tracking a simple trajectory. High tracking weight was set in the MPC objective function to minimise tracking error, reject the human disturbance, and ensure similar realised trajectories for each trial with all RMS tracking errors below 0.0835 rad. During the trials, EMG signals, $\textrm{ch}_1$ and $\textrm{ch}_2$, and exoskeleton torque input, $\tau_e$, were recorded over numerous cycles. Using the relaxed condition (R) as a baseline for zero human torque, reference human torque, $\tau_{hr}$, was calculated for the four non-relaxed effort conditions (C) using superposition of the exoskeleton torque profiles, where
\begin{equation}
    \tau_{hr} = \tau_{e,R} - \tau_{e,C}.
\end{equation}
The SEA used for these experiments uses high resolution encoders to measure rotational deflection of the series elastic component, enabling highly accurate measurement of exoskeleton torque, $\tau_e$, at the knee joint. Linear regression was applied across the involvement condition (C) reference torque profiles for each subject to find the optimal parameters for (\ref{eqn:torque}) shown in Table \ref{tab:tau_h_est}. RMS values of torque estimation error
\begin{equation}
\text{RMSE}_{\hat{\tau}_h} = \sqrt{\frac{1}{n}\sum(\hat{\tau}_h-\tau_{hr})^2}
\end{equation}
where $n=60000$ is the total number of data samples across six repetitions of 4 s trials in the five human involvement conditions sampling at 500 Hz. The reference human torque $\tau_{hr}$ and estimated human torque $\hat{\tau}_h$ for all human involvement conditions over a sample 4 s period are shown in Fig. \ref{fig:tau_hs}. Satisfied agreement from the linear agonist-antagonist model can be seen. The estimation accuracy (mean RMSE$_{\hat{\tau}_h}$ = 2.82 Nm across all three subjects), is sufficient to ensure an accurate assistance mode detection using the FLA described in Section \ref{sec:fla}. The normalized RMS torque estimation error, NRMSE$_{\hat{\tau}_h}$ is calculated by scaling RMSE$_{\hat{\tau}_h}$ by the range of reference human torque, $(\tau_{hr,\textrm{max}}-\tau_{hr,\textrm{min}}) = 40.6$~Nm, giving NRMSE$_{\hat{\tau}_h}$ of 0.0695 and accuracy of~93.05\%. This accuracy is comparable to previous studies involving knee joint torque estimation \cite{Chen2017}.

\subsection{Experimental Protocol}
Performance of the proposed MPC approach with FLA for assistance mode detection is evaluated for knee joint flexion/extension movements in a seated position. Knee joint movements of the swinging leg during a gait cycle have a pendulum oscillatory flexion/extension behaviour that is close to the natural walking frequency \cite{Huo2018}. This makes such exercise suitable for early-stage gait rehabilitation in stroke patients.

For the target movement profiles, we considered a sinusoidal knee joint angle trajectory with amplitude of 1 rad. This is defined by
\begin{equation} \label{eqn:trajectory}
    \theta_r(t) = 0.5\cos(2\pi ft-\pi)+0.7
\end{equation}
where $\theta_r$ is the reference joint angle in radians, $f$ is repetition frequency in Hz, and $t$ is time in seconds. For the tests in this paper $f=0.25$, meaning one exercise repetition lasts 4 s. The frequency is chosen to closely match that of knee joint movement in normal walking gait \cite{Xu2019a}. Trials were conducted over short periods of time (24 s) for each human involvement condition in order to prevent subject fatigue and discomfort.

The aim was to demonstrate tracking performance of the proposed AAN-MPC controller and the adaptation capability for different human involvement conditions, i.e., adjusting the nature of human-robot interaction. The controller is tested operating in the following modes:

\subsubsection{Passive Mode} The subjects are relaxed, contributing negligible torque, relying on the exoskeleton to generate torque in order to successfully complete the tracking task. This is a robot-dominant condition.

\subsubsection{Active-Assist Mode} The subjects voluntarily generate joint torque to track the target extension/flexion movements. The exoskeleton should show compliance and only provide the deficient torque required to successfully complete the rehabilitation task. This mode enables human-robot cooperation. Therefore this condition is neither human nor robot-dominant. The extent to which human/robot is responsible for generating torque for the rehabilitation task is dictated by the assist mode penalty parameter, $p_A = 0.5$, configured within the FLA, as described in Section \ref{sec:fla}.

\subsubsection{Safety Mode} The subjects generate significant joint torque which obstructs the target extension/flexion movement. In practice this could be due to stroke patient pathology, limited range of motion, or subject discomfort for example. This is a potentially dangerous human-dominant condition in which the subject is not cooperating. For in-clinic and in-home assessment and rehabilitation, the safety of human–robot interactions, specifically patient–robot interaction could be a major
concern \cite{Atashzar2017}. It is highly beneficial for the controller to identify this and adapt appropriately. The exoskeleton should show high levels of compliance and not prioritise generating torque for the trajectory tracking task.
\section{Experimental Results} \label{sec_results}

\begin{figure*}[t]
\centering
\includegraphics[width=18cm]{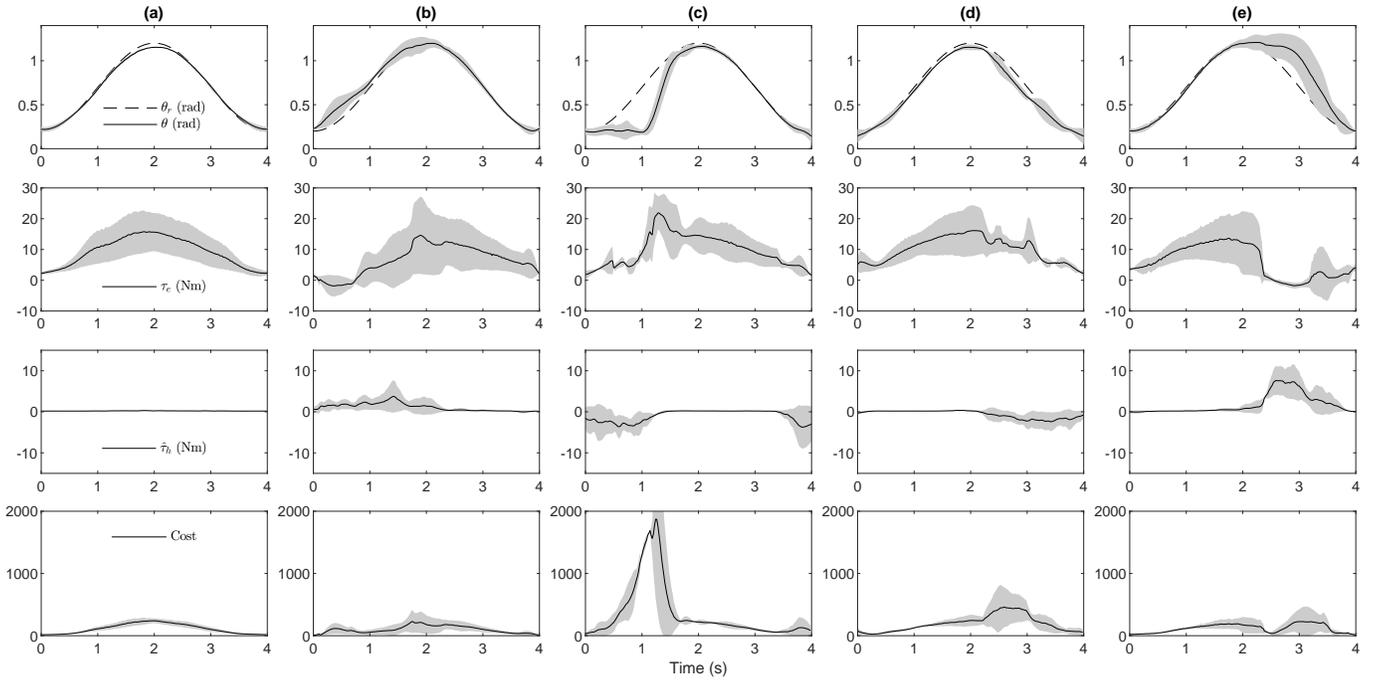}
\caption{Tracking results from all three subjects for sinusoidal reference of frequency 0.25 Hz for MPC-AAN controller with FLA assistance mode detection in five human involvement conditions. The plots show the average values and shaded standard deviation (SD). Arranged from left to right in columns: a) relaxed (R) b) extension assist (EA) c) extension resist (ER) d) flexion assist (FA) and e) flexion resist (FR). From top to bottom, rows contain reference and actual trajectory, $\theta_r$ and $\theta$, robot torque, $\tau_e$, estimated human torque, $\hat{\tau}_h$, and objective function cost.}
\label{fig:tracking_error_bars}
\end{figure*}

The AAN controller results are presented for all subjects performing a knee joint tracking task in five human involvement conditions: relaxed (R), extension assist (EA), extension resist (ER), flexion assist (FA), and flexion resist (FR). R should invoke passive mode, EA and FA should invoke active-assist mode, and ER and FR should invoke safety mode.  

Performance is evaluated in terms of suitable robot compliance (reflected in tracking error), human-robot cooperation/interaction (reflected in overall human torque ratio), and suitable assistance mode detection during extension (E) and flexion (F) movement phases of the exercise.

For evaluation of human-robot interaction, human torque ratio, $R_h$, is defined as
\begin{equation}
    R_h = \frac{\Bar{\hat{\tau}}_h}{\Bar{\hat{\tau}}_h+\Bar{\tau}_e}
\end{equation}
where $\Bar{\hat{\tau}}_h$ and $\Bar{{\tau}}_e$ denote the root-mean-square (RMS) of the estimated human joint torque and the exoskeleton torque, respectively. 

\begin{figure}
    \centering
    \includegraphics[width=\columnwidth]{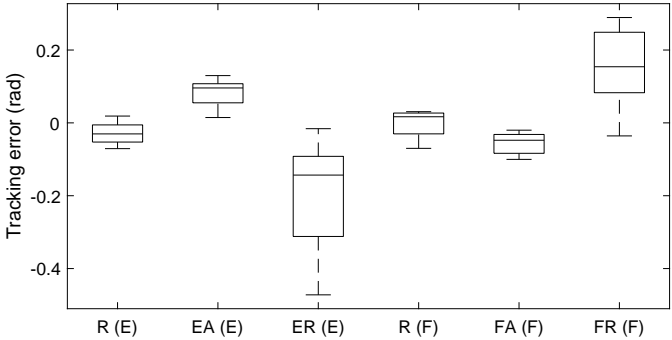}
    \caption{Box plots comparing percentiles of tracking error (rad) for extension (E), from 0 to 2 s, and flexion (F), from 2 to 4 s, in the five human involvement conditions (R, EA, ER, FA, and FR) averaged across all subjects. Large error values reflect robot compliance.}
    \label{fig:boxes}
\end{figure}

\subsection{Trajectory Tracking}

Tracking results averaged across all three subjects for a sinusoidal reference trajectory of frequency 0.25 Hz can be seen for the five human involvement conditions in Fig. \ref{fig:tracking_error_bars}, with standard deviation (SD) shown as a shaded region. The significant SD seen in exoskeleton torque across all involvement conditions is a result of varying subject properties, with smaller and lighter subjects requiring lower exoskeleton torque to complete the exercise. SD is also present in the human torque estimation, $\hat{\tau}_h$, due to subjects applying different levels of effort for varying time during the trials.

In the robot-dominant condition with human relaxed (R) the controller operates in passive mode. It closely follows the target trajectory with very low tracking error, as shown in Fig. \ref{fig:boxes} for both extension and flexion (medians of -0.03 rad and 0.02 rad). There is minimal tracking SD across the relaxed trials. Low human torque is correctly estimated in this condition with minimal SD. The total objective function cost remains low and very little compliance is shown by the robot. This is reflected in very close trajectory tracking, despite a lack of human cooperation, with torque contributions very low for extension ($R_h=0.036$), and flexion ($R_h=0.042$).

During EA and FA involvement conditions the FLA recognises assistive human involvement, sets $\mu_A$ high, and sets the controller to active-assist mode. In this mode it correctly permits greater tracking error. Fig. \ref{fig:tracking_error_bars}(b) and Fig. \ref{fig:tracking_error_bars}(d) show the controller behaving with greater compliance as the subjects begin to apply voluntary torque. This is shown in Fig. \ref{fig:boxes} as significantly increased tracking error, with median errors of 0.10 rad and -0.05 rad during the extension and flexion phases respectively. These is a significant increase from the corresponding phases for R condition.

During ER and FR conditions, the controller shows even greater compliance, as instructed by setting the safety mode penalty $p_S = 1.0 > p_A = 0.5$. This compliance can clearly be seen in Fig. \ref{fig:tracking_error_bars}(c) and Fig. \ref{fig:tracking_error_bars}(e) where the joint angle significantly deviates from the reference trajectory. This leads to much increased overall cost, caused by significant tracking error (medians of -0.14 rad and 0.15 rad for extension and flexion phases respectively). For the ER condition, once the FLA no longer observes high $\mu_S$ and detects human relaxation, the tracking error is minimised with high corrective torque, on average greater than 20 Nm around 1.25 s, reducing observed cost. The augmented Lagrangian optimization method adjoins constraint equations to the overall MPC objective function. Cost levels as high as 2000 are shown for ER due to activation of a maximum velocity constraint during trajectory recovery.

\subsection{Human-Robot Interaction}

\begin{figure}[t]
    \centering
    \includegraphics[width=\columnwidth]{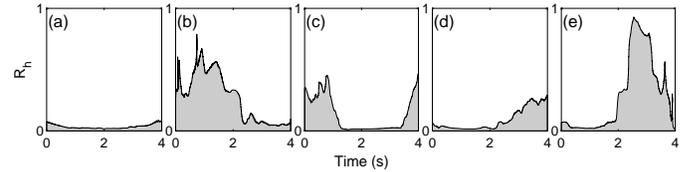}
    \caption{Human torque ratio, $R_h$, throughout the exercise tasks, averaged across all three subjects for the five human involvement conditions: a) relaxed (R), b) extension assist (EA), c) extension resist (ER), d) flexion assist (FA), and e) flexion resist (FR).}
    \label{fig:human_ratio}
\end{figure}

The average human torque ratio, $R_h$, throughout the exercise (from 0 to 4 s) across all subjects for each condition is shown is Fig. \ref{fig:human_ratio}. In R involvement condition, very little human cooperation is shown as human torque ratio is very low ($R_h=0.036$ for extension and $R_h=0.042$ for flexion).

For EA and FA, increased human-robot cooperation is expected, and corresponding human torque ratio is observed, with $R_h=0.46$ for the extension phase of EA and $R_h=0.18$ for the flexion phase of FA. The human effort is shown in Fig. \ref{fig:tracking_error_bars}(b) during EA, between 0.5 s and 2 s, as exoskeleton torque, $\tau_e$, is much lower compared to R condition, despite noticeable tracking error.

For ER and FR, the calculated human torque ratio values are significantly greater ($R_h=0.23$ and $R_h=0.59$ for extension and flexion phases respectively) compared to R, however the human effort observed during these phases obstructs the exercise, so does not signify human-robot cooperation. Instead the human torque is impeding the robot and is not helpful for the tracking task.

\begin{figure}
    \centering
    \includegraphics[width=\columnwidth]{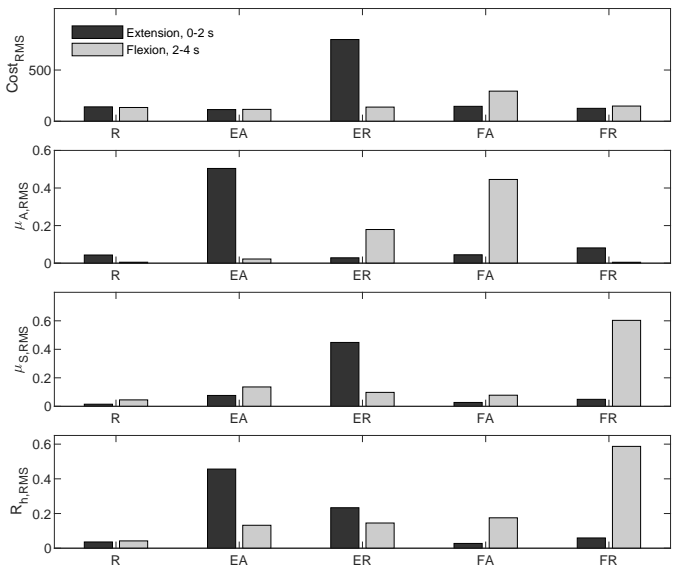}
    \caption{Bar charts showing, from top to bottom, RMS values of cost, active-assist mode likelihood, $\mu_A$, safety mode likelihood, $\mu_S$, and human torque ratio, $R_h$, for all subjects. Results are divided into extension phase in dark grey, from 0 to 2 s in the trials, and flexion phase in light grey, from 2 to 4 s for relaxed (R), extension assist (EA), extension resist (ER), flexion assist (FA), and flexion resist (FR) human involvement conditions.}
    \label{fig:bars}
\end{figure}

\subsection{Assistance Mode Detection with FLA}

\begin{figure*}
    \centering
    \includegraphics[width=18cm]{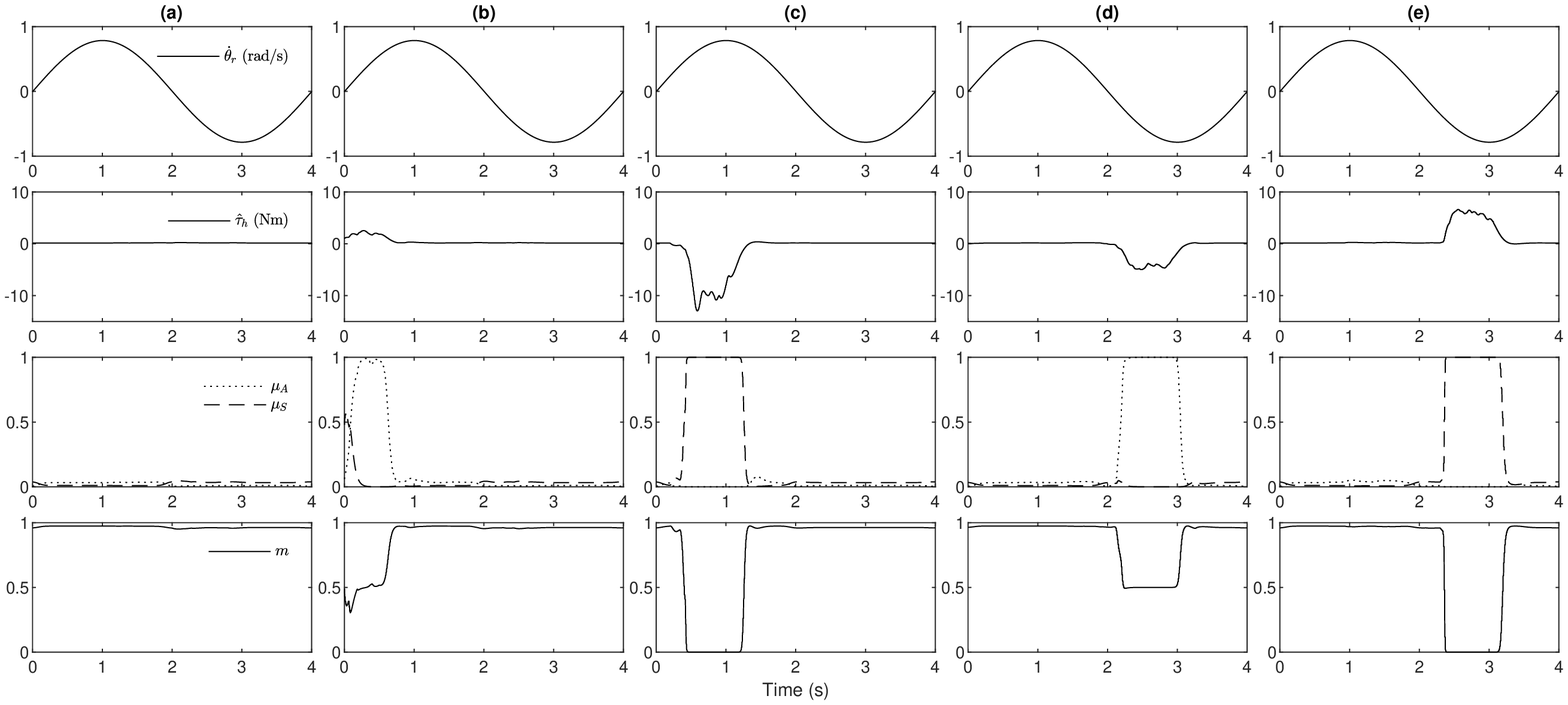}
    \caption{Fuzzy logic algorithm results for subject 1 trajectory tracking trials for MPC-AAN controller with FLA assistance mode detection in five human involvement conditions. Arranged from left to right in columns: a) relaxed (R) b) extension assist (EA) c) extension resist (ER) d) flexion assist (FA) and e) flexion resist (FR). From top to bottom, rows are FLA inputs of reference joint angular velocity $\dot{\theta}_r$ and estimated human torque $\hat{\tau}_h$, active-assist and safety mode likelihoods, $\mu_A$ and $\mu_S$, and assistance mode variable $m$.}
\label{fig:fla}
\end{figure*}

The RMS assistance mode likelihood values, $\mu_A$ and $\mu_S$, averaged across all three subjects are shown for the five human involvement conditions in Fig. \ref{fig:bars}. The values show that the FLA successfully identifies the correct mode of operation for relaxed, human-assistive and human-resistive involvement conditions. The greatest RMS active-assist mode likelihood values, $\mu_A$, occur during the extension and flexion phases of EA and FA respectively (0.50 and 0.45) and the greatest RMS safety mode likelihood values, $\mu_S$, occur during the extension and flexion phases of ER and FR respectively (0.45 and 0.60).

The FLA input signals, $\dot{\theta}_r$ and $\hat{\tau}_h$, active-assist and safety mode likelihood values, $\mu_A$ and $\mu_S$, and the assistance mode variable, $m$, for each human involvement condition are shown for subject 1 in Fig. \ref{fig:fla}. The FLA performs well, consistently identifying the appropriate assistance mode at any point during the task. For R condition, both $\mu_A$ and $\mu_S$ remain low, as the FLA recognises human relaxation and the controller to operates in passive mode. The assistance mode function, $m$, behaves as expected, falling closer to zero for safety mode than for active-assist mode, given $p_S = 1.0 > p_A = 0.5$. $\mu_A$ and $\mu_S$ are activated on a mutually exclusive basis.
\section{Discussion and Conclusion} \label{sec_conclusion}

In this paper, an AAN controller is proposed which combines a constrained nonlinear MPC approach with a FLA for assistance mode detection. Human torque is estimated from processed EMG signals using a linear agonist-antagonist torque model. Although the EMG-based method is dependent on muscle activity measurement to extract the torque output, the method benefits from advanced prediction of human behaviour. This gives an advantage over sensorless approaches such as nonlinear disturbance observers. Another advantage is distinction of disturbance caused by human effort from other external forces. The estimated human torque is used by the MPC for deriving the robot torque required to successfully follow a target trajectory. It is also used by the FLA for evaluating human involvement condition and selecting the most suitable assistance mode. Human-robot interaction ranges on a spectrum with human-assisting at one extreme, human-relaxed in the middle, and human-resisting at the other extreme. For convenient analysis, human involvement conditions are discretely defined for the knee joint tracking task as relaxed (R), extension assist (EA), extension resist (ER), flexion assist (FA), and flexion resist (FR). The FLA enables the controller to change behaviour fluidly, operating in assistance modes that fall on a continuous spectrum. Without human torque, the controller operates in passive mode, which is a servo-like robot-dominant condition. This is achieved by setting high tracking weight, $w_\theta$, in the MPC cost function. In this mode the exoskeleton generates nearly all of the required torque for the task. In active-assist mode, the controller shows compliance and enables robot-human cooperation, contributing less torque to the exercise task, with the subject applying greater effort. In safety mode, the controller exhibits very high levels of compliance for the safety and comfort of the human, and trajectory tracking is given low priority, with very little robot torque produced for the task. Moreover, to further improve the safety, a reference trajectory adaptation algorithm (see Appendix \ref{sec:adaptation}) can be used to halt the reference trajectory when the likelihood of the safety mode is higher than a given threshold. The trajectory adaption can prevent the subject experiencing relatively long-term uncomfortable levels of resistance to the rehabilitation task. It can also prevent the subject falling behind with the task, enabling them to safely pick up at the reference point where they were last comfortable and remain engaged with the exercise.

The controller is demonstrated tracking a 0.25 Hz sinusoidal reference trajectory in the five representative human involvement conditions. The FLA is shown to identify the appropriate assistance mode during each of them. A trajectory adaptation method is proposed as a potential utilisation of high level human involvement information from the FLA. The human resistance limit for trajectory halting can be defined specific to each subject for a particular task and the trajectory is shown adapting to let the subject relax and catch up, with inferred human resistance falling back below the limit before the task continues.

For the recovery of motor function, active patient participation in generating a target exercise movement is highly important. The proposed controller enables this without the need for repetitive training trials which can use up valuable time in clinical situations.

In future work, tests will be conducted involving more general and complex trajectories. The controller will also be applied to a more advanced scenario with a multiple DOF system model, such as upright walking or squatting exercises. Clinical testing with survivors of stroke or elderly subjects with mobility problems is a long-term goal. Routines for automatically adjusting controller behaviour parameters $p_A$ and $R_{th}$ will be investigated, optimising these parameters online, to make the controller better deal with subject-specific range of movement and fatigue. The model-based nature of the controller makes it highly adaptable with relatively little parameter tuning, so it is applicable to almost any lower limb rehabilitation device, given a representative system model can be derived. We will also investigate situations with intricate muscle group involvements and consider implementing a more sophisticated muscle model for human torque estimation.

\appendices
	
\section{Trajectory Adaptation} \label{sec:adaptation}

Trajectory adaptation, triggered by excessive human resistance detected by the FLA, is investigated as an optional feature, and shown as
\begin{equation}
    \theta_r(k+1) =
  \begin{cases}
                         0.5\cos(2\pi ft_{k+1}-\pi)+0.7 & \text{if $\mu_S \leq R_{th}$} \\
                                   \theta_r(k) & \text{if $\mu_S > R_{th}$}
  \end{cases}
\end{equation}
where $k$ denotes the $k$th sample. $R_{th}$ is a human resistance threshold. 

The time-dependent trajectory (\ref{eqn:trajectory}) is halted when the likelihood of the safety mode is higher than a given threshold, $\mu_S > R_{th}$. The reference joint angle remains constant until the estimated likelihood of the safety mode, $\mu_S$, falls back below the threshold, at which point the reference trajectory continues. The threshold $R_{th}$ value can be selected based on the subject's range of movement, fatigue level, or other clinical factor and can be updated with an online algorithm. 

\printbibliography

\end{document}